\documentclass{article}
\usepackage{spconf,amsmath,graphicx,hyperref}
\usepackage{cite}
\usepackage{svg}

\usepackage{amsmath,amssymb,amsfonts}
\usepackage{algorithmic}
\usepackage{graphicx}
\usepackage{textcomp}
\usepackage{xcolor}

\title{Enhanced Graph Neural Networks using K-Hop Gaussian Diffusion}
\name{Xuling Zhang$^{1}$, Peng Wang$^{1}$, Daiyan Li$^{1,2}$, Aoran Huang$^{1}$, Zeiwei Chen$^{1}$, Yongkui Yang$^{1\ast}$
\thanks{Equal contribution: X. Zhang, P. Wang, and D. Li contributed equally as first authors; A. Huang and Z. Chen contributed equally as second authors. 
$^{\ast}$Corresponding author: Yongkui Yang, yk.yang@siat.ac.cn.}
}
\address{$^{1}$Shenzhen Institutes of Advanced Technology, Chinese Academy of Sciences, Shenzhen, China\\
$^{2}$Southern University of Science and Technology, Shenzhen, China}

\begin{document}
%
\maketitle
\begin{abstract}

Most graph neural network (GNN) cores rely on graph convolutions, typically implemented as message passing between direct (single-hop) neighbors. In many real-world graphs, edges can be noisy or poorly defined, limiting information propagation to local neighborhoods. Existing diffusion kernels, such as Personalized PageRank (PPR) and Heat Kernel, alleviate this issue through global propagation, but still struggle with complex local structures and distant node noise. To address these limitations, we propose a K-Hop Gaussian (KHG) diffusion kernel as a preprocessing module for graph data. KHG introduces multi-hop diffusion with Gaussian weighting for remote nodes, balancing local and global information propagation before applying standard GNNs. Experiments on multiple benchmark datasets demonstrate that KHG significantly outperforms traditional message-passing GNNs, as well as PPR and Heat Kernel diffusion, particularly in noisy or structurally complex graphs.
\end{abstract}

\section{Introduction}

Graph neural networks (GNNs) have become an important tool for analyzing graph-structured data, with applications in social networks, transportation networks, molecular graphs \cite{3}, biological networks, financial transaction networks \cite{5}, and citation graphs \cite{6}. By integrating deep learning with graph data, GNNs achieve strong performance in node classification, graph classification, and link prediction. Most GNNs are built upon message passing neural networks (MPNNs), which update node embeddings by aggregating neighborhood information. Representative models include GCN \cite{9}, SGC \cite{2}, ChebNet \cite{8}, TAGCN \cite{11}, and JKNet \cite{7}, covering techniques from spectral convolution to skip-connection aggregation. While deep neural networks generally benefit from more layers, GNNs suffer from over-smoothing when the number of layers is large \cite{li2020}, with optimal performance often observed within 2--4 hops \cite{Klicpera2019}. Techniques such as DropEdge \cite{rong2019} and DropNode, inspired by dropout \cite{srivastava2014}, attempt to mitigate this issue, but random edge/node removal may damage graph structure. Another challenge is the information bottleneck \cite{alon2020}, where long-range dependencies are poorly captured. Moreover, many GNNs assume simple unweighted, undirected graphs, overlooking richer structural patterns such as self-loops and multi-edges. To address these limitations, we propose the \textbf{K-Hop Gaussian (KHG) diffusion kernel}, which introduces multi-hop diffusion with Gaussian weighting. This design balances local and global propagation, suppresses noise from distant nodes, and improves robustness compared with existing diffusion kernels such as Personalized PageRank (PPR) \cite{Klicpera2019} and the Heat Kernel.  

\textbf{Our main contributions are threefold:} 
(1) We propose the KHG diffusion kernel to mitigate over-smoothing and information bottlenecks in deep GNNs. (2) KHG integrates multi-hop diffusion via Gaussian weighting and is a modular, plug-and-play preprocessing component for existing GNNs. (3) Extensive experiments on node and graph classification benchmarks demonstrate its superiority over PPR and Heat kernels.

\begin{figure*}[t]
\centering
\includegraphics[width=0.6\textwidth]{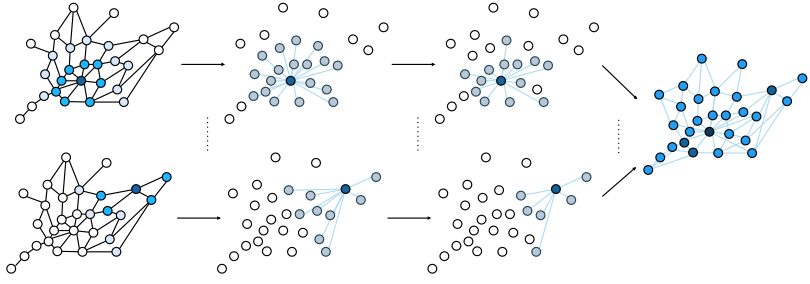}
\caption{Illustration of the K-Hop Gaussian Diffusion process with $K=2$. The dark blue node denotes the diffusion source, blue nodes correspond to the 1-hop neighborhood, and light blue nodes represent the 2-hop neighborhood. After diffusion, edges are ranked by their diffusion weights, and only the top-weighted edges are retained. This procedure is repeated for all nodes, and the resulting diffusion graphs are finally merged to construct the K-Hop Gaussian Diffusion graph.}
\label{fig:kgdc_process}
\vspace{-3mm}
\end{figure*}
\section{RELATED WORK}
\label{sec:RELATED WORK}
\textbf{Different GNN and Message Passing:}  
In GNNs, classical models such as GCN \cite{9}, SGC \cite{2}, and ChebNet \cite{8} aggregate only 1-hop neighbors, limiting long-range modeling, while excessive multi-hop diffusion leads to over-smoothing. GraphSAGE \cite{hamilton2017inductive} is a full GNN model that aggregates information from 1-hop and 2-hop neighborhoods to balance computational cost and over-smoothing. Its hop is typically restricted to 2 because larger neighborhoods can introduce noise and increase model complexity. In contrast, our K-Hop Gaussian (KHG) diffusion is a preprocessing module rather than a full GNN, which allows flexible selection of larger K-hop ranges to propagate information while controlling contribution from distant nodes via Gaussian weighting. Diffusion kernels generalize this idea: PPR \cite{Klicpera2019} uses personalized random walks, and the Heat Kernel models heat flow, both lacking explicit hop-wise control. GRAND \cite{Chamberlain2021grand} improves flexibility but requires iterative or sampling steps. KHG achieves efficient, noise-robust, and deterministic multi-hop diffusion in closed form. \textbf{Diffusion Mechanism:}  
In GNNs, information propagation is key to learning node relationships. Classical models such as GCN \cite{9}, SGC \cite{2}, and ChebNet \cite{8} rely on 1-hop aggregation, limiting long-range modeling, while excessive multi-hop diffusion leads to over-smoothing. Methods like PPR \cite{Klicpera2019} and the Heat Kernel capture broader context but lack explicit hop-wise control. GRAND \cite{Chamberlain2021grand} enhance flexibility via learnable or continuous diffusion, yet incur iterative or sampling overhead. Our K-Hop Gaussian (KHG) diffusion provides a closed-form alternative: a Gaussian-decayed multi-hop kernel that smoothly balances local and global propagation while remaining efficient and noise-robust. \textbf{Different GNN and Message Passing:}Gaussian filtering is a classic denoising technique in image processing, also widely used in deep vision models \cite{he2016}. Inspired by spectral filtering on graphs \cite{bruna2014}, we extend it to GNNs to control multi-hop propagation.  In images, the Gaussian kernel for a pixel $(x,y)$ is $G(x,y) = \frac{1}{2\pi\sigma^2}\exp\!\left(-\frac{x^2+y^2}{2\sigma^2}\right),$
where $\sigma$ controls the smoothing scale. Large $\sigma$ yields broad filtering, while small $\sigma$ preserves fine details. Analogously, in graphs we replace spatial distance with hop distance $i$, yielding the weight $w(i,\sigma) = \exp\!\left(-\frac{i^2}{2\sigma^2}\right)$,which decays smoothly with $i$. This bridges local and global propagation: small $\sigma$ enforces local smoothing, while larger $\sigma$ incorporates broader context, mitigating over-smoothing in deep GNNs.  

\begin{figure*}[t]
\centering
\includegraphics[width=0.66\textwidth]{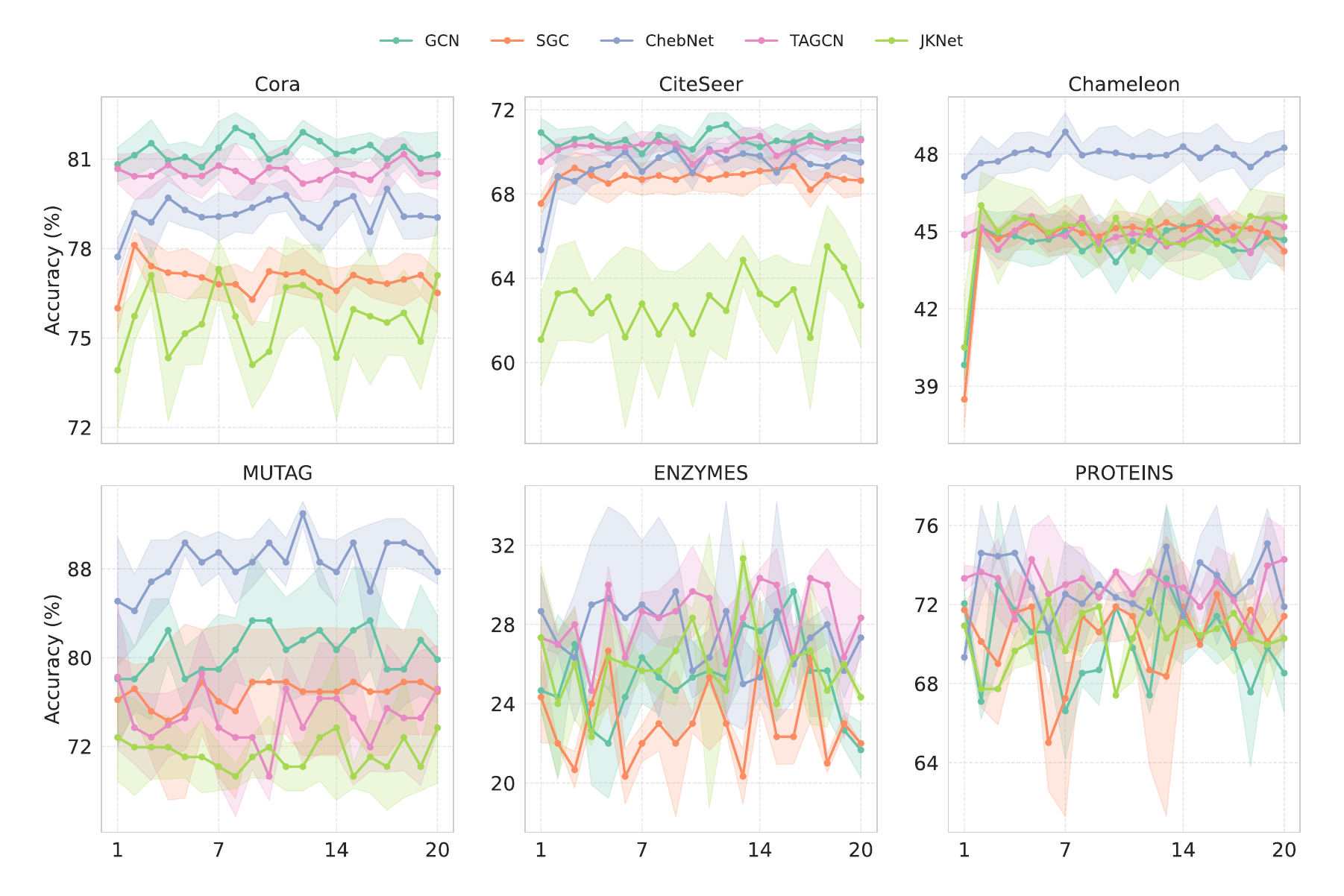}
\vspace{-3mm}
\caption{K-Hop Gaussian Diffusion Model Performance Comparison.}
\label{fig}
\end{figure*}

\section{METHOD}
\label{sec:METHOD}
\subsection{Graph Normalization and Transition Matrix}

K-Hop Gaussian (KHG) Diffusion is a Gaussian kernel-based method that enables multi-hop propagation in GNNs, ensuring that information decays smoothly with distance to mitigate over-smoothing and noise. The input is the adjacency matrix $A \in \mathbb{R}^{N \times N}$, where $A_{ij}=1$ if an edge exists between nodes $i$ and $j$, and $0$ otherwise. The goal is to construct a diffusion matrix that balances local smoothness and long-range dependencies. A key step is normalizing $A$ to obtain a stable transition matrix $T$. Without normalization, degree imbalance can cause uneven propagation. Two widely used forms are: \textbf{Symmetric normalization}: $T_{\mathrm{sym}} = D^{-\tfrac{1}{2}} A D^{-\tfrac{1}{2}}$, where $D_{ii} = \sum_j A_{ij}$. This form balances both incoming and outgoing degrees, as in GCN \cite{9}. \textbf{Column normalization}: $T_{\mathrm{col}} = D^{-1}A$, which distributes a node’s information evenly among its neighbors and prevents excessive dilution for low-degree nodes. Both $T_{\mathrm{sym}}$ and $T_{\mathrm{col}}$ serve as the base operator for multi-hop propagation ($T^k$), later combined with Gaussian weights in KHG diffusion. Compared to unnormalized $A$, these normalized forms guarantee stability and preserve structural balance in heterogeneous graphs.

\subsection{K-Hop Gaussian Diffusion Mechanism}

Let $A\in\mathbb{R}^{N\times N}$ be the adjacency matrix and $D=\mathrm{diag}(d_i)$ the degree matrix. We first form a normalized transition operator $T$ (we use either symmetric or column normalization): $T_{\mathrm{sym}} = D^{-1/2} A D^{-1/2},  T_{\mathrm{col}} = D^{-1} A.$
The $i$-hop transition operator is the $i$-th power of $T$ and satisfies the recursion
\begin{equation}
T^{(1)} = T,\qquad T^{(i)} = T^{(i-1)}T \quad (i\ge 2),
\end{equation}
or element-wise $T^{(i)}_{uv} = \sum_{k} T_{uk}\,T^{(i-1)}_{kv}$,
which encodes reachability and aggregated propagation within $i$ hops.
To control hop-wise influence we introduce a Gaussian hop-weight
\begin{equation}
w(i,\sigma)=\exp\!\left(-\frac{i^2}{2\sigma^2}\right),\qquad i=1,\dots,K,
\end{equation}
where $\sigma>0$ is the scale and $K$ the maximal hop cutoff. The unnormalized multi-hop kernel aggregates weighted powers: $\widetilde{D}_K \;=\; \sum_{i=1}^K w(i,\sigma)\,T^{(i)}.$
Normalizing by the scalar weight sum $Z(\sigma)=\sum_{i=1}^K w(i,\sigma),$
we obtain the final diffusion kernel
\begin{equation}\label{eq:DK}
D_K \;=\; \frac{\widetilde{D}_K}{Z(\sigma)} \;=\; \frac{\sum_{i=1}^K w(i,\sigma)\,T^{(i)}}{\sum_{i=1}^K w(i,\sigma)}.
\end{equation}
If $T=U\Lambda U^{-1}$ (orthogonal $U$ when $T$ is symmetric), then
\[
T^{(i)}=U\Lambda^{i}U^{-1},\qquad
\widetilde{D}_K = U\Big(\sum_{i=1}^K w(i,\sigma)\Lambda^{i}\Big)U^{-1},
\]
so $D_K$ acts as a polynomial spectral filter
\begin{equation}
D_K = U\,\mathrm{diag}\!\big(g(\lambda_j)\big)\,U^{-1},\qquad
g(\lambda)=\frac{\sum_{i=1}^K w(i,\sigma)\lambda^{i}}{\sum_{i=1}^K w(i,\sigma)}.
\end{equation}
Thus KHG is a degree-$K$ polynomial filter with Gaussian coefficients. For comparison, PPR corresponds to geometric weights $D_{\mathrm{PPR}}=\alpha\sum_{i\ge0}(1-\alpha)^i T^i=\alpha(I-(1-\alpha)T)^{-1}$ and heat diffusion to factorial/exponential series $\exp(tT)=\sum_{i\ge0}t^i/i!\,T^i$. Building $D_K$ explicitly may densify; instead compute $D_KX$ for a feature matrix $X\in\mathbb{R}^{N\times d}$ via recursion. Let $Y^{(0)}=X$ and $Y^{(i)} = T\,Y^{(i-1)},\quad i=1,\dots,K,$ then
\begin{equation}
D_K X \;=\; \frac{\sum_{i=1}^K w(i,\sigma)\,Y^{(i)}}{\sum_{i=1}^K w(i,\sigma)}.
\end{equation}
This avoids storing $T^{(i)}$ and leads to the simple iterative procedure: iterate $K$ sparse multiplications $Y\leftarrow T Y$, accumulate $w(i,\sigma)Y$ and divide by $Z(\sigma)$. when implemented at feature-level (sparse $T$). Explicit matrix construction can densify to $O(N^2)$ storage in the worst case. In practice use sparse accumulation and optional top-$m$ row-wise sparsification to keep $D_K$ sparse. Parameter semantics are intuitive: small $\sigma$ concentrates mass on low-order powers (local smoothing), large $\sigma$ spreads mass across hops (more global aggregation); $K$ limits maximal propagation depth.

\section{Experiment}

\begin{table*}[h]
\caption{Performance Comparison of GNN Models with Different Diffusion Mechanisms across Datasets} 
\label{dataset-table}

\centering
\footnotesize
\setlength{\tabcolsep}{0.1pt} 
\begin{tabular*}{0.97\textwidth}{l@{\extracolsep{\fill}}lccccccc} 
\hline
 
\textbf{Model} & \textbf{Diffusion} & \textbf{Cora} & \textbf{CiteSeer} & \textbf{PubMed} & \textbf{Chameleon} & \textbf{MUTAG} & \textbf{ENZYMES} & \textbf{PROTEINS} \\
\hline
GRAND 
& -l    & 81.21±0.52 & 70.95±0.61 & 78.10±0.50 & 45.00±0.80 & 82.10±3.20 & 28.90±0.75 & 72.40±1.05 \\
& -nl   & 81.55±0.48 & 71.05±0.44 & 78.25±0.55 & 45.20±0.73 & 82.85±2.90 & 29.10±0.81 & 72.95±1.20 \\
& -nl-rw& 81.70±0.61 & 71.20±0.39 & 78.40±0.62 & 45.10±0.69 & 83.00±3.10 & 29.25±0.65 & 73.00±0.95 \\

\hline
GCN & None   & 80.82±0.57 & 70.92±0.68 & 77.55±0.30 & 39.82±0.62 & 78.07±6.05 & 24.67±0.92 & 72.04±0.66 \\
    & PPR    & 81.30±1.04 & 70.34±0.68 & 78.28±0.40 & 45.79±1.46 & 74.74±2.77 & 24.33±1.57 & 68.31±1.31 \\
    & Heat   & 81.24±1.17 & 71.08±1.51 & 77.50±0.33 & \textbf{46.49±2.20} & 82.74±4.47 & 26.27±0.60 & 70.03±0.96 \\
    & KHG    & \textbf{82.04±0.51 (K=8)} & \textbf{71.30±0.56 (K=11)} & \textbf{78.55±0.76 (K=14)} & 45.26±1.02 (K=14) & \textbf{83.33±6.05 (K=9)} & \textbf{29.67±0.46 (K=16)} & \textbf{73.32±3.76 (K=3)} \\
\hline
SGC & None   & 76.00±0.83 & 67.55±0.46 & 70.79±0.11 & 38.49±1.10 & 76.19±3.92 & 24.33±2.31 & 71.73±0.22 \\
    & PPR    & 77.36±0.57 & 69.04±0.56 & 71.57±0.25 & 44.93±0.61 & 77.37±4.28 & 24.53±2.17 & 67.73±1.78 \\
    & Heat   & 77.50±1.25 & \textbf{69.58±0.39} & \textbf{72.79±0.35} & 44.72±1.08 & 77.01±3.13 & 24.34±1.81 & 70.35±0.93 \\
    & KHG    & \textbf{78.11±0.42 (K=2)} & 69.32±0.81 (K=3) & 71.70±0.27 (K=5) & \textbf{45.35±0.37 (K=13)} & \textbf{77.81±4.74 (K=6)} & \textbf{26.67±2.77 (K=14)} & \textbf{72.52±1.33 (K=16)} \\
\hline
ChebNet & None & 77.72±0.63 & 65.54±1.50 & 76.75±0.37 & 47.13±0.66 & 85.09±5.75 & 28.67±1.85 & 69.33±0.44 \\
        & PPR  & 79.78±1.50 & 69.74±1.23 & 77.23±0.26 & 44.20±0.56 & 76.32±2.92 & 28.27±2.97 & 73.74±0.65 \\
        & Heat & 79.96±1.13 & 69.92±1.04 & 77.24±0.20 & 44.87±1.93 & 82.63±1.85 & 28.13±1.86 & 74.38±1.05 \\
        & KHG  & \textbf{80.01±0.57 (K=17)} & \textbf{70.12±0.34 (K=9)} & \textbf{77.35±0.20 (K=20)} & \textbf{48.86±0.75 (K=17)} & \textbf{92.98±1.09 (K=12)} & \textbf{29.67±2.31 (K=9)} & \textbf{75.08±1.77 (K=19)} \\
\hline
TAGCN & None & 80.67±0.49 & 69.57±0.59 & 78.03±0.34 & 44.87±0.67 & 78.25±6.52 & 27.33±3.23 & 73.32±0.66 \\
      & PPR  & 81.07±0.96 & 70.16±1.28 & 78.88±0.25 & 45.06±1.81 & 78.05±2.69 & 27.87±4.70 & 72.91±0.54 \\
      & Heat & 81.14±1.26 & 70.61±0.30 & 77.97±0.20 & 45.40±2.34 & 78.16±5.38 & 28.14±1.40 & 72.92±1.40 \\
      & KHG  & \textbf{81.16±0.54 (K=19)} & \textbf{70.75±0.42 (K=14)} & \textbf{78.95±0.69 (K=18)} & \textbf{45.57±0.79 (K=16)} & \textbf{78.57±3.92 (K=6)} & \textbf{30.33±0.46 (K=14)} & \textbf{74.28±1.55 (K=20)} \\
\hline
JKNet & None & 73.92±1.93 & 61.09±2.24 & 74.60±1.21 & 40.50±2.03 & 72.81±3.92 & 27.33±3.70 & 70.93±0.00 \\
      & PPR  & 77.04±1.94 & 62.36±1.79 & 75.29±1.15 & 43.77±1.33 & 73.16±3.97 & 27.43±1.01 & 70.16±1.47 \\
      & Heat & 75.54±2.13 & 64.01±2.02 & 74.63±1.44 & 45.39±2.25 & 73.26±2.35 & 27.13±1.01 & 71.82±0.94 \\
      & KHG  & \textbf{77.31±0.99 (K=7)} & \textbf{65.51±1.94 (K=18)} & \textbf{78.80±0.55 (K=19)} & \textbf{46.01±1.29 (K=2)} & \textbf{73.68±6.52 (K=14)} & \textbf{31.33±0.92 (K=13)} & \textbf{72.20±2.21 (K=12)} \\
\hline
\end{tabular*}
\end{table*}

\subsection{Dataset and Baselines}
In our experiments, we evaluated the proposed K-Hop Gaussian (KHG) diffusion against a range of representative GNN baselines, including GCN \cite{9}, SGC \cite{2}, ChebNet \cite{8}, TAGCN \cite{11}, JKNet \cite{7}, as well as advanced diffusion methods GRAND \cite{Chamberlain2021grand}. For node classification, we used three benchmark citation networks: Cora , CiteSeer, Pubmed\cite{namata2012query} and Chameleon \cite{rozemberczki2021}. For graph classification, we considered three molecular and protein datasets: ENZYMES \cite{borgwardt2005}, MUTAG \cite{debnath1991}, and PROTEINS \cite{dobson2003}. Datasets without official splits (e.g., Chameleon, ENZYMES, MUTAG, PROTEINS) were randomly divided into training/validation/test sets in a 7:2:1 ratio using a fixed random seed of 42 to ensure reproducibility. Each experiment was repeated multiple times, and results are reported with 95\% confidence intervals to ensure stability and robustness.

\subsection{Performance Comparison}

Table~I reports results of four diffusion mechanisms (None, PPR, Heat, KHG) on three node classification datasets (Cora, CiteSeer, Chameleon) and three graph classification datasets (MUTAG, ENZYMES, PROTEINS) combined with five GNN models (GCN, SGC, ChebNet, TAGCN, JKNet). Overall, KHG consistently outperforms other kernels. On Cora, for instance, GCN+KHG achieved 82.0\%, surpassing both None (80.8\%) and PPR/Heat. On Chameleon, ChebNet+KHG reached 48.9\%, 3.7\% higher than None. In graph classification, the improvement is even larger: ChebNet+KHG achieved 93.0\% on MUTAG, 7.9\% above None, and GCN+KHG gained 5.0\% on ENZYMES. These results demonstrate that Gaussian weighting effectively enhances robustness in both sparse and complex graphs. The choice of $K$ strongly influences performance. For smaller, sparse graphs (Cora, MUTAG), small $K$ (\(<10\)) is optimal, while for denser or heterogeneous graphs (Chameleon, ENZYMES) moderate $K$ (\(>10\)) yields better accuracy by capturing more global information without excessive noise. This trend highlights KHG’s adaptability to different graph scales and structures.
\begin{figure}[t]
\centering
\includegraphics[width=0.4\textwidth]{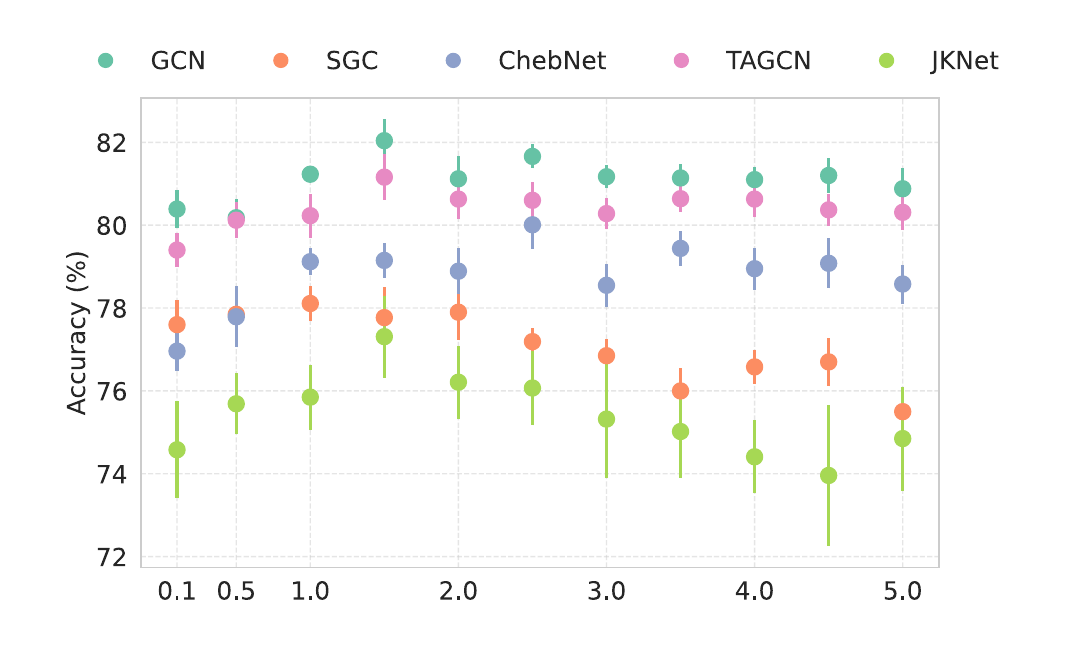}
\caption{Effect of $\sigma$ on accuracy (Cora dataset, fixed $K$).}
\label{fig:sigma}
\end{figure}

\textbf{Efficiency.}  
Unlike PPR and Heat, which rely on iterative power series or matrix exponentiation, KHG is a closed-form preprocessing step. On PubMed (19k nodes, 88k edges, 500-dimensional features), constructing $D_KX$ with $K=14$ takes about $1.8$s and 180MB memory. In contrast, PPR with $\alpha=0.1$ requires $\sim$6.5s due to iterative convergence, while Heat needs to store multiple dense expansions, consuming over 400MB. This demonstrates that KHG achieves comparable or better accuracy with significantly lower computational overhead, making it scalable to large graphs.

\textbf{Influence of $\sigma$ on KHG.}  
As shown in Fig.~\ref{fig:sigma}, small $\sigma$ values cause rapid weight decay, suppressing distant information and leading to poor accuracy. As $\sigma$ increases, performance improves since remote neighbors contribute more. Most models achieve the best results when $\sigma\in[1.0,3.0]$, with JKNet and TAGCN peaking around $\sigma=1.5$, showing that Gaussian weighting balances local and global propagation.

\section{CONCLUSION}
This paper proposes a new K-Hop Gaussian (KHG) diffusion mechanism, which is a data preprocessing method for graphs. Aiming to solve the shortcomings of existing graph neural networks (GNN) in processing complex graph structures and noisy data. By introducing Gaussian-weighted multi-hop diffusion, this method effectively balances the propagation of local and global information and avoids excessive smoothing and information bottleneck problems in deep networks. The KHG diffusion kernel can not only suppress the noise impact of distant nodes but also enhance the feature retention of low-degree nodes, thereby improving the performance of the overall model. Our experiments show that this method significantly improves the performance of baseline GNNs. Future work will be devoted to developing more efficient diffusion algorithms to apply this method on larger-scale graphs and explore its potential in other graph learning tasks.

\section{Acknowledgment}
This work was supported in part by National Natural Science Foundation of China (Grant No. U25A2054), in part by Shenzhen Science and Technology Projects (Grant No. JSGGZD20220822095602005).

\bibliographystyle{IEEEbib}
\bibliography{strings,refs}

\end{document}